\documentclass[runningheads]{llncs}
\usepackage{ntheorem}
\usepackage{graphicx}
\usepackage[%
    font={small,sf},
    labelfont=bf,
    format=hang,    
    format=plain,
    margin=0pt,
    width=0.8\textwidth,
]{caption}
\usepackage[list=true]{subcaption}
\theoremstyle{remark}

\newcommand{\xdownarrow}[1]{%
  {\left\downarrow\vbox to #1{}\right.\kern-\nulldelimiterspace}
}
\begin{document}
\title{Toward Privacy and Utility Preserving Image Representation}
%
%\titlerunning{Abbreviated paper title}
% If the paper title is too long for the running head, you can set
% an abbreviated paper title here
%
\author{Ahmadreza Mosallanezhad \and
Yasin N. Silva \and
Michelle V. Mancenido \and Huan Liu}
\authorrunning{Mosallanezhad et al.}
% First names are abbreviated in the running head.
% If there are more than two authors, 'et al.' is used.
%
\institute{Arizona State University, Tempe AZ, USA 
\email{\{amosalla,ysilva,mmanceni,huanliu\}@asu.edu}}
\maketitle              % typeset the header of the contribution
\begin{abstract}
Face images are rich data items that are useful and can easily be collected in many applications, such as in $1$-to-$1$ face verification tasks in the domain of security and surveillance systems. Multiple methods have been proposed to protect an individual's privacy by perturbing the images to remove traces of identifiable information, such as gender or race. However, significantly less attention has been given to the problem of protecting images while maintaining optimal task utility. In this paper, we study the novel problem of creating privacy-preserving image representations with respect to a given utility task by proposing a principled framework called the Adversarial Image Anonymizer (AIA). AIA first creates an image representation using a generative model, then enhances the learned image representations using adversarial learning to preserve privacy and utility for a given task. Experiments were conducted on a publicly available data set to demonstrate the effectiveness of AIA as a privacy-preserving mechanism for face images.

\end{abstract}
\section{Introduction}
Security and surveillance systems, such as those found in private industries (e.g., biometric access control systems) and public domains (e.g., face recognition systems at airports and traffic thruways), acquire images of people's faces for verification and identification tasks. The ease of collecting data on private citizens raises concerns about violating privacy-preserving contracts or expectations~\cite{pri}, because organizations have been known to exercise their prerogative to sell information on individuals or have been subject to malicious attacks that compromised users' privacy~\cite{mosallanezhad2019deep,beigi2020privacy}. For instance, in 2019, a malicious cyber-attack to a US Customs and Border Protection subcontractor exposed traveler's photos \cite{guardian}. Due to such threats to individual liberties and privacy, one method that has been proposed to protect an individual's private information is to anonymize it before sharing. While some recent studies have shown that face images contain user-related private information such as gender or race~\cite{img1,img2,img3}, research on protecting face images from adversarial attacks has been limited ~\cite{vgan}. 
%For instance, in 2006, AOL's search data leak resulted in performing re-identification tasks on search logs and queries \cite{pass}.

Two general approaches have been proposed for preserving privacy on image data. The first method, known as visual privacy, perturbs images so that a human cannot infer a user's private attributes. Text representations have many hidden attributes which can be used for different sentiment analysis tasks~\cite{Heidari-etal-2020-ELMO,Heidari-etal-2020-Machinelearning,Heidari-etal-2020-Social_bots}. Similar to the text representations, image representations can also have both useful and sensitive attributes. Thus, in order to protect images, the second method creates a representation from the image data~\cite{vgan}, which then replaces the original image in image-based applications. An advantage of visual privacy is that the perturbed images can easily be used in various image-based tasks. This approach, however, does not provide the same level of privacy-preserving effectiveness as the second method~\cite{vgan}. Furthermore, current methods do not guarantee that the perturbed image is still useful in a specific utility task.

In this paper, we address the challenge of creating a privacy-preserving image representation while simultaneously preserving the image's utility for a given image-base task. To address this challenge, we propose the AIA (\underline{A}dversarial \underline{I}mage \underline{A}nonymizer) framework composed of three main modules: (1) a component to encode images for representation learning; (2) a component for privacy-preserving representation learning; and (3) a component to preserve the utility of the learned representations. The main contributions of this paper are (1) the study of the novel problem of joint privacy and utility preserving image representation; (2) a principled framework (AIA) that integrates adversarial learning and a generative model to create a privacy-preserving image representations which can be used with a given utility task; and (3) extensive experiments on a publicly available data set to demonstrate the effectiveness of AIA for creating both privacy preserving image representations for a specific utility task.

% The rest of the paper is organized as follows: Section 2 presents related work in the area, while Section 3 presents the general problem statement. We then introduce the proposed AIA framework in Section 4 and demonstrate the performance of this proposed approach in Section 5. We provide conclusions and future directions in Section 6.

\section{Related Work}
This section briefly describes relevant and related work on the following domains: (1) image generation and representation;  and (2) adversarial learning on generative models.

\noindent
\textit{\textbf{Image Generation and Representation.}} Most of the previous work on image generation focuses on using generative models for image generation. For example, Cheung et al. proposed a method for generating images using auto-encoders by studying the hidden and latent factors in image representations~\cite{hiddenfact}. Mathieu et al. proposed the use of adversarial training to create an image representation that can be used to generate sharp looking images~\cite{vardeep}. More recently, Tran et al. proposed a face recognition model based on Generative Adversarial Networks (GAN) ~\cite{vardeep2}. Despite not addressing privacy preservation, these contributions showed that latent factors in image representations could provide important information about the images such that one can generate the original images using the image representations. Some of these previous results were recently leveraged in the work of Chen et al. to build a model for privacy-preserving facial expression recognition through GAN-based image representation learning. This model is capable of generating privacy-preserving images~\cite{vgan}. A disadvantage of this model, however, is that the generated images are human recognizable.

% \subsection{Adversarial Learning on Generative Models}
\noindent
\textit{\textbf{Adversarial Learning on Generative Models.}} Generative Adversarial Networks (GAN) are neural networks that use adversarial training in a game environment to train two networks, a generator and a discriminator~\cite{GAN}. This method can generate high-quality images and image representations. Despite being extensively used, GANs could present some problems such as stability issues and mode collapse~\cite{GANProb}. Makhzani et al. proposed the use of adversarial auto-encoders (GAN-based probabilistic auto-encoders) to create better image representations in order to generate high-quality images ~\cite{AAE}.

Our work is distinct from previous efforts in two ways. First, we consider generating a privacy-preserving image representation instead of generating an actual image. Second, we simultaneously optimize for both privacy and a given utility task during the creation of image representations. Our model can be particularly useful preserving privacy in systems that are designed for a specific task, e.g. 1-to-1 face matching.

\section{Problem Statement} \vspace{-5pt}
Let $\mathcal{X} = \{ x_1, x_2, ..., x_N \}$ denote a set of $N$ grey-scaled images where each image $x_i$ is composed of a matrix $x_i \in {\rm I\!R}^{1\times n\times m}$. Let $p$ denote a private attribute that users do not want to disclose such as gender. We address the following problem:
\begin{problem}
Given a set of images $\mathcal{X}$ and private attribute $p$, learn an anonymizer $f$ that can learn an image representation $\mathbf{g}\in {\rm I\!R}^{1 \times k}$ such that: (1) [Privacy preservation] an adversary cannot infer the targeted user's private attribute $p$, and (2) [Utility preservation] the image representation $\mathbf{g}$ can be effectively used in a given task $\mathcal{T}$ such as 1-to-1 face matching. The problem can be expressed as:
 
\begin{equation}
    \mathbf{g_i} = f(x_i, p, \mathcal{T})
\end{equation}

\end{problem}

Due to the success of auto-encoders in learning image representations~\cite{AAE}, we use auto-encoders with adversarial training to create both utility and privacy preserving image representations in this paper. 

\section{Proposed Method}\vspace{-5pt}
The main components of the Adversarial Image Anonymizer (AIA) framework appear in  Figure~\ref{fig:AIA}. The input to the system is a grey-scale image, while the output is a privacy-preserved vector $\mathbf{g}\in {\rm I\!R}^{1 \times k}$. The model has $3$ main components: (1) an auto-encoder composed of an encoder that learns the input image representation and a decoder that can reconstruct the input image given its representation, (2) an adversary which tries to infer the user's private attribute using the image representation, and (3) a task $\mathcal{T}$ that is used to preserve the utility of the image representation. In this framework, we first train each component individually and, then, use an adversarial loss function to enhance the overall model. We present the details of our proposed method next.

\subsection{Learning Image Representations}
The goal of this component is to generate an image representation that can be used in different tasks. The image representations in our model are created using an auto-encoder. This is a generative model that is trained in an unsupervised way to learn latent representations of the input images. A key feature of auto-encoders is their dimensionality reduction ability which was an important reason to integrate it instead of using GANs. Another reason of using an auto-encoder instead of a GAN is to prevent some of the issues in GAN-based models such as stability problems and time-consuming training ~\cite{GAN2}. The auto-encoder consists of the following components: 

\begin{figure}[t]\vspace{-10pt}
    \centering
    \includegraphics[scale=0.15]{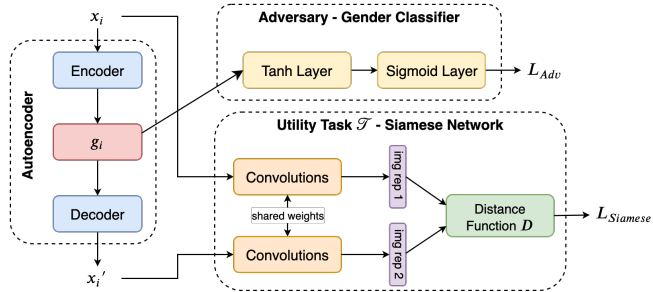}
    \caption{Architecture of the Adversarial Image Anonymizer (AIA).}
    \label{fig:AIA}
\end{figure}

% \begin{itemize}
\noindent
\textit{\textbf{Encoder.}} The encoder learns a latent representation of the input image and aims at reducing its dimensionality. $x_i$ is a grey-scaled image with dimensions $n\times m$. To create a representation of the image we use Convolutional Neural Networks (CNN) to learn filters that can identify the important parts of an input image. In our model, we first use a convolution layer $Conv$ to create feature maps from the input image. Then, we apply an activation function $F$ on the gathered features and use a pooling layer $P$. The pooling layer helps to select the parts of the features with strong correlation to the input image. We refer to the output of these three layers as a block $L_j^{enc}$ where $j$ represents the $j^{th}$ encoder block:

    \begin{equation}
        L_j^{enc} = P( F( Conv(x) ) )
    \end{equation}
    
    We use a stack of these blocks to create an image representation. Observe that in order to convert the output of the final pooling layer to a vector $\mathbf{g}$, we use a flattening layer that converts a matrix into a vector:
    
    \begin{equation}
        \mathbf{g} = Flatten(L_{J}^{enc})
    \end{equation}
    
    where $J$ is the index of the final encoder's layer.
    
\noindent    
\textit{\textbf{Decoder.}} The decoder tries to reconstruct the input image using the previously generated representation. The process of decoding the image representation $\mathbf{g}$ is similar to the one in the encoding phase, but in reverse. First, we use a convolution layer to create feature maps, then we apply an activation function to the features. As for the final layer, instead of using a pooling layer which generates a smaller matrix, we use an up-sampling layer $U$ to increase the size of the reconstructed image:
  
    \begin{equation}
        L_l^{dec} = U( F ( Conv(\mathbf{g}) ) )
    \end{equation}
    
    where $L_l^{dec}$ indicates the $l^{th}$ decoder block. The final output of the decoder is the reconstructed image $\mathbf{x'}$ which is then used to train the auto-encoder.
% \end{itemize}

After creating the auto-encoder, we train it using the input image $x_i$ and the reconstructed image $x'_i$. This will result in learning a representation vector $\mathbf{g}$ which is expected to capture useful information of the input image.

\subsection{Adversarial Training}
Creating an image representation using an auto-encoder alone could result in privacy issues~\cite{vgan}. For example, a well-trained gender classifier could predict the gender of a person using the corresponding image representation. To prevent this issue, we integrate adversarial training to create privacy-preserving image representations. In this component, we use a powerful adversary, i.e. gender classifier, to further improve the learned representation of the auto-encoder. Because our goal is to create a privacy-preserving representation for a given task $\mathcal{T}$, we use the loss value of this task as a penalty to generate private learned representation.

\subsection{Optimization Algorithm}
The training process in the proposed model consists of two parts. In the first part, we train each component (auto-encoder, adversary, and the given task $\mathcal{T}$) separately:
\begin{itemize}
    \item \textit{Auto-encoder}: For the encoder component, we use stacked CNN blocks, each containing a 2D convolution layer with leaky ReLU as the activation function; and an average pooling which operates on the output of the activation function. The decoder has also two stacked CNN blocks. Each block has a convolution layer with a leaky ReLU activation function. The output of the activation function goes through an up-sampling layer to create an output image with similar size to the input image. We train the auto-encoder using the Binary Cross Entropy loss function between the input image $x_i$ and the reconstructed image $x'_i$. This loss function calculates how well the auto-encoder has predicted the image:    
    \begin{equation}
        L_{AE} = x' \cdot \log x + (1 - x') \cdot \log(1 - x)
    \end{equation}

    \item \textit{Adversary}: In our model, we use a high-quality gender classifier as the adversary. The adversary acquires an image representation $\mathbf{g}$ as input and predicts whether the image corresponds to a female or male face. We use a three-layer neural network to output gender probablity $\mathbf{o}$:
    \begin{equation}
        \mathbf{o} =  sigmoid(\mathbf{W_A^{(2)}} (\tanh(\mathbf{W_A^{(1)}} \mathbf{g} + \mathbf{b_A^{(1)}} )) + \mathbf{b_A^{(2)}})
    \end{equation}
    \iffalse
    \begin{equation}
        r = sigmoid(\mathbf{W_A^{(2)}} \mathbf{o} + \mathbf{b_A^{(2)}} )
    \end{equation}   
    \fi
    where $\mathbf{W_A^{(.)}}$, $\mathbf{b_A^{(.)}}$, are learnable weights. This classifier is also trained using the same Binary Cross Entropy (BCE) loss $L_{Adv}$.

    \item \textit{Task $\mathcal{T}$}: In this paper we consider a 1-to-1 face matching task, which verifies if two input images are of the same individual. We use the well-known Siamese network for this task which acquires two images $x_i$ and $x_i'$ as input and returns their representations. The distance of the two representations is then calculated based on the similarity between $x_i$ and $x_i'$. We train this model using the following loss function:
    \begin{equation}
        L_{Siamese} = (1 - y) \frac{1}{2} D^2 + y \frac{1}{2} max(0, m - D)^2
    \end{equation}
    where $D$ as the Euclidean distance and $m$ a constant margin.

\end{itemize}

\textit{Adversarial Training:} After training each component separately, we use the following loss function to enhance the autoencoder for generating privacy and utility preserving representations based on the feedback from the utility and the adversary components:
\begin{equation}
    L_{Total} = L_{AE} + \alpha L_{Siamese} - \beta L_{Adv}
\end{equation}
where $\alpha$ and $\beta$ indicate the contribution of each loss value. In our model, we use a powerful attacker to ensure privacy even from other unseen attackers. In our experiments we show that our model can preserve privacy of user's private attributes from different attackers.

\section{Experiments} \vspace{-5pt}
We performed multiple experiments to evaluate the performance of the proposed model. We aim to answer the following questions: (\textbf{Q1}) how well does our method protect users' private attribute, i.e., gender?; (\textbf{Q2}) how well does our method preserve the utility of an image with respect to a given task $\mathcal{T}$?; and (\textbf{Q3}) what is the relation between privacy and utility? To answer \textbf{Q1}, we use an adversary to test if it can detect gender based on the perturbed representations. For \textbf{Q2}, we study the performance of the utility task before and after perturbing the learned image representations. Finally, to answer \textbf{Q3}, we study how the effectiveness of preserving users' privacy impacts the utility of the learned representations.

\subsection{Data}
In this study, we use two different publicly-available datasets, CelebA~\cite{liu2015faceattributes} and VGG face datasets~\cite{parkhi2015deep}. CelebA consists of over 200K celebrity images with various metadata~\cite{liu2015faceattributes}. %We use images alongside with the gender attribute as their private information to train and test our model. 
VGG consists of 2,622 identities where each identity has different images~\cite{parkhi2015deep}. In both datasets, we use gender as the private attribute. % Moreover, this dataset is categorized into male and female which we consider this attribute as private information. 

\subsection{AIA Implementation Details}
The adversary is a gender classifier which has $6$ convolution layers with 32, 64, 64, 128, 128, and 256 channels, respectively. After the convolution layers, we use a one-layer neural classifier with Softmax on the output layer. The auto-encoder is composed of an encoder and a decoder. The encoder has two convolution layers with $8$ and $12$ channels, while the decoder contains three convolution layers with $256, 128$, and $1$ channels. The final convolution layer in the decoder converts the $128$ features into $1$ feature to generate the image corresponding to the associated input image. Finally, for the utility task $\mathcal{T}$, we use a Siamese network. This network has three convolution layers and three fully connected layers after flattening the output of the convolution layers. The three convolution layers have $4, 8$ and $8$ channels, respectively. The fully connected network has an input layer, a hidden layer with $500$ neurons, and an output layer with $128$ neurons. The output of this network is an embedded image representation which is then used to calculate the distance between two input images.

\subsection{Experimental Design}  \vspace{-5pt}
We use the following baseline methods for comparision:

\begin{itemize}
    \item \textbf{\textsc{Original}:} this method does not use any anonymization and only outputs the learned image representation from the auto-encoder's output. 
    \item \textbf{Random:} this method randomly changes $50\%$ of each learned image representation to make it private. Because each value in the image representation is within $[0.0, 1.0]$, we randomly select $50\%$ of the numbers and change each to a random number sampled from $[0.0, 1.0]$.
    \item \textbf{AIA\textbackslash T:} this is a modified version of our proposed method that does not use the adversarial training for preserving the utility of the learned image representation.
\end{itemize}

\subsection{Experimental Results} \vspace{-5pt}

In this subsection, we evaluate AIA's performance using a gender classifier (adversary) and a similarity task (utility).

\textbf{Privacy (Q1)}. Table~\ref{table:results} compares the accuracy of the different methods for gender-detection and the utility task. In the Gender Privacy column, lower accuracy indicates that the gender classifier for that method was less effective at predicting a user's gender based on the image representations. We observe that AIA is better at protecting privacy than the original approach. While AIA cannot preserve users' privacy as well as AIA\textbackslash{}T and the random methods, it preserves utility significantly better than these methods. The random method, despite being more effective at privacy preservation, generated significantly lower utility, implying that it generated useless representations.

\begin{table}[ht!]\vspace{-10pt}
\centering
\caption{Accuracy of the private-attribute classifier and the utility task. In these results, $\alpha = \beta = 0.5$.}
\begin{tabular}{l|c|c|c|c}
 & \multicolumn{2}{c|}{CelebA Dataset} & \multicolumn{2}{c}{VGG Dataset} \\\hline
Method & Privacy ($\downarrow$ better) & Utility ($\uparrow$ better) &  Privacy ($\downarrow$ better) & Utility ($\uparrow$ better) \\ \hline
Original                    & \%78.01 & \%88.87 & \%81.21 & \%91.31 \\
Random                      & \%52.12 & \%56.89 & \%51.34 & \%53.67 \\
AIA\textbackslash{}T        & \%62.53 & \%69.34 & \%65.67 & \%66.29 \\ \hline
AIA                         & \textbf{\%64.96} & \textbf{\%78.64} & \textbf{\%68.13} & \textbf{\%77.96}
\end{tabular}
\label{table:results}
\end{table}

\begin{figure}[t!]\vspace{-15pt}
\centering
\subcaptionbox{Gender Inference ($\downarrow$ better)}{\includegraphics[width=0.50\textwidth]{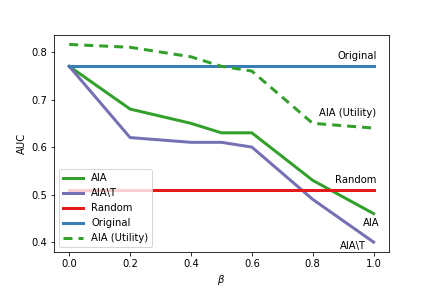}}%
\hfill
\subcaptionbox{Utility ($\uparrow$ better)}{\includegraphics[width=0.50\textwidth]{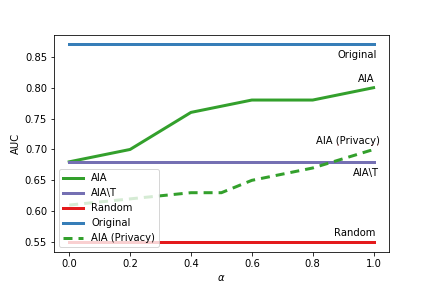}}%
\caption{AUC scores for private-attribute inference and utility tasks for different values of $\alpha$ and $\beta$.} % Lower AUC values in the gender inference task represents higher privacy levels while higher AUC values in the utility task indicate higher utility. The dashed lines show the AUC of the other task (utility for gender inference and privacy for the utility task).}
\label{fig:results}
\end{figure}

\textbf{Utility (Q2)}. As illustrated in Table~\ref{table:results}, our method performs better than the Random and AIA\textbackslash{}T methods for preserving the utility of the image representations. While the Original method has the highest accuracy level, it provides the worst privacy-preserving guarantees. This highlights the need for changing image representations in order to preserve users' privacy. Changing image representations randomly will perverse users' privacy but it will also greatly decrease their utility. While AIA\textbackslash{}T is relatively effective preserving privacy, it lacks the utility benefits that AIA provides. This is because AIA\textbackslash{}T does not have any component that forces the auto-encoder to preserve the utility of image representations.

\textbf{Privacy Utility Trade-off (Q3)}. 
AIA has two main parameters, $\beta$ controls the contribution of the gender classifier, while $\alpha$ controls the contribution of the utility task. In Figure~\ref{fig:results} we show the performance of our model and the baselines for different values of $\alpha$ and $\beta$ using the CelebA dataset. For each parameter, we hold one of them as $0.5$ and vary the other one from 0 to 1. As shown in this figure, achieving higher levels of utility with AIA results in lower levels of privacy assurance and vice versa. Figure 2.a shows AIA's utility level using a dashed line for different values of $\beta$ while $\alpha=0.5$. We can observe that using a $\beta$ value larger than $\alpha$ could result in substantial utility loss. Figure 2.b shows the inverse. The dashed line in this figure shows the gender inference AUC values (where higher values correspond to lower privacy) for different values of $\alpha$ while $\beta=0.5$. These results indicate that reaching higher levels of utility can result in significant privacy loss. The problem of utility and privacy preservation is consequently a multi-objective, trade-off optimization problem where each objective antagonizes the other. From Figure~\ref{fig:results}, we conclude that using similar values for $\alpha$ and $\beta$ provides a reasonable balance between both privacy and utility. In general, a judicious choice for the two tuning parameters depends on the application domain. If privacy is more important than utility, $\beta$ should be higher than $\alpha$, and vice versa. 

\subsection{Visualization}
\begin{figure}[t!]\vspace{-10pt}
    \centering
    \includegraphics[scale=0.8]{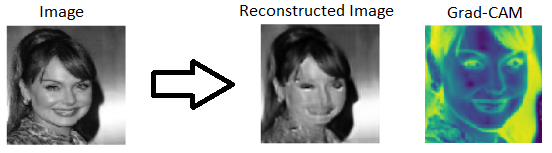}
    \caption{Visualization results. We used the decoder to reconstruct the image and used Grad-CAM to visualize the learned features from the gender classifier.}
    \label{fig:viz}
\end{figure}

To gain additional insights, we visualized and analyzed the learned image representations using a low value of $\alpha=0.2$. We used Grad-CAM~\cite{selvaraju2017grad}, a well-known CNN visualization technique. Given a label and an image, this method generates an activation map that can be used to identify the areas of the image that are most relevant to the label. Figure~\ref{fig:viz} shows both the reconstructed and Grad-CAM images for a sample photo and the result of the gender classifier. The reconstructed image in Figure~\ref{fig:viz} shows the outcome of the anonymization process. We used the decoder component from the auto-encoder to reconstruct the image using its representation. The Grad-CAM image shows the important parts of the image that resulted in the classifier predicting a female label. We can observe that the classifier focuses on the hair length and the eyes, as well as the location of the cheekbones and the shape of the chin. The adversarial training in AIA influences the auto-encoder to hide these pieces of information that result in more accurate gender labels. \vspace{-2pt}

\section{Conclusions and Future Work}\vspace{-3pt}
Protecting the privacy of citizens has been a widespread concern in an era where the intentional or unintentional propagation of private information has been an unfortunate byproduct of machine learning. We recognized that limited attention has been given to mechanisms that simultaneously preserve an individual's privacy, while preserving the intended utility of a machine or deep learning model. Thus, we proposed the AIA framework that uses an auto-encoder to learn image representations and then enhances these representations using adversarial training. Initial results showed an interesting trade-off between utility and privacy which results in outcomes that offer better privacy while having only a small impact on utility. The performance results showed that AIA performed well overall in comparison to the baseline methods. An immediate future work direction is the extension of the proposed framework to consider multiple private attributes. This work also has salient applications in the area of bias in deep learning models, a direction that is currently being pursued by the authors. % Currently, we are performing more experiments to control for margins of errors on utility and privacy measures. An obvious future direction is to extend this work in the presence of multiple private attributes. This work also has salient applications in the area of bias in deep learning models, a direction that is currently being pursued by the authors. 
% Initial results showed an interesting trade-off between utility and privacy, which results in solutions that are sub-optimal for one of the two objectives. Despite this, AIA performed well overall in comparison to other baseline methods. 

\section{Acknowledgements}\vspace{-3pt}
This material is based upon work supported by the U.S. Department of Homeland Security under Grant Award Number 17STQAC00001-04-00.\footnote{Disclaimer: “The views and conclusions contained in this document are those of the authors and should not be interpreted as necessarily representing the official policies, either expressed or implied, of the U.S. Department of Homeland Security.}

% REFERENCES
\bibliography{samplepaper}
\bibliographystyle{splncs04}

\iffalse

\fi
\end{document}